\documentclass{article}

\PassOptionsToPackage{numbers, compress}{natbib}
\usepackage[final]{neurips_2021}

\usepackage[utf8]{inputenc} 
\usepackage[T1]{fontenc}    
\usepackage{hyperref}       
\usepackage{url}            
\usepackage{booktabs}       
\usepackage{amsfonts}       
\usepackage{nicefrac}       
\usepackage{microtype}      
\usepackage{xcolor}         
\usepackage{graphicx}
\usepackage{subcaption}
\usepackage{amssymb}
\usepackage{mathtools}
\usepackage{amsmath}
\usepackage{comment}
\usepackage{url}
\usepackage{natbib}
\usepackage{wrapfig}

\usepackage{algorithmic}
\usepackage[ruled]{algorithm2e}
\usepackage{multirow}
\usepackage{eqparbox,array}

\usepackage{makecell}
\newcommand{\norm}[1]{\left\lVert#1\right\rVert}

\newlength{\maxwidth}
\newcommand{\algalign}[2]
{\makebox[\maxwidth][r]{$#1{}$}${}#2$}

\makeatletter
\newcommand{\printfnsymbol}[1]{%
  \textsuperscript{\@fnsymbol{#1}}%
}
\makeatother

\title{A Probabilistic Framework for Knowledge Graph Data Augmentation}

\author{%
  Jatin Chauhan\thanks{equal contribution} \hspace{0.1mm} \thanks{corresponding author is currently at Google AI} \\
  IIT Hyderabad \\
  \texttt{chauhanjatin100@gmail.com} \\
   \And
   Priyanshu Gupta\printfnsymbol{1}\thanks{corresponding author is currently at Microsoft Research} \\
  IIT Kanpur \\
   \texttt{priyanshu.42g@gmail.com} \\
   \AND
   Pasquale Minervini \\
   University College London \\
  \texttt{p.minervini@gmail.com} \\
}
\SetKwComment{Comment}{$\triangleright~$}{}

\SetKwInOut{Input}{Input}
\SetKwInOut{Output}{Output}
\begin{document}

\maketitle

\begin{abstract}
We present NNMFAug, a probabilistic framework to perform data augmentation for the task of knowledge graph completion to counter the problem of data scarcity, which can enhance the learning process of neural link predictors. Our method can generate potentially diverse triples with the advantage of being efficient and scalable as well as agnostic to the choice of the link prediction model and dataset used. Experiments and analysis done on popular models and benchmarks show that NNMFAug can bring notable improvements over the baselines.

\end{abstract}

\section{Introduction}
The most widely used representation of \textit{Knowledge Bases (KBs)} is in the form of \textit{Knowledge Graphs (KGs)} where the nodes represent entities that are connected by relations in form of a directed graph. Extensive research in the past decade has shown that these KGs can be extremely useful for many core NLP tasks such as relation extraction \cite{mintz-etal-2009-distant, vashishth-etal-2018-reside}, summarization \cite{DBLP:journals/corr/abs-2005-01159}, question answering \cite{DBLP:journals/corr/BordesWU14}, dialog systems \cite{ma_dialog}, recommender systems \cite{zhang_rec_kg} and many more due to their simplistic structure and the ability to abstract out facts and knowledge.

Despite their success, a major drawback of KGs is their incompleteness \cite{10.1145/2623330.2623623}. Since the actual number of valid KG triples can be extremely large, ensuring that they are complete can be a daunting task, if done manually. This can in-turn stagnate the improvements on downstream tasks. The task of \textit{Knowledge Graph Completion} \cite{NIPS2013_1cecc7a7} extensively focuses on tackling this issue by learning models, commonly known as \textit{link predictors}, that can complete any triple with partial information. More recently, neural network based methods \cite{10.5555/3045390.3045609, DBLP:journals/corr/DettmersMSR17, sun2018rotate, balazevic-etal-2019-tucker}, commonly referred to as \textit{neural link predictors}, have become the state of the art for KG completion task. However, since these models are supervised learners, their ability is directly tied to the amount of training data available.

Recent threads of research present empirical and theoretical arguments to suggest that data augmentation can improve the performance of deep learning models \cite{10.1145/3361242.3361259, chen2020grouptheoretic} by non-trivial margins while also leading to improved generalization. \cite{volpi2018generalizing, Hoffer_2020_CVPR}.
Inspired from these works, we propose \textit{NNMFAug}, a novel method to perform \textbf{data augmentation} over knowledge graphs  to improve the performance of neural link predictors. 

We develop a \textit{probabilistic} framework that is \textit{agnostic} to the choice of link prediction model and dataset from which new and diverse triples can be sampled while ensuring scalability and efficiency. Further, we present a new training routine to gradually increase the number of these newly sampled triples that are augmented to the training set as a function of the training epochs completed. Experiments and analysis done on popular neural link predictors and benchmarks show that our technique can bring notable improvements over the baselines trained on the available training data only.

\section{Method}
We first provide the formulation of the probabilistic framework that fits a distribution over the set of all possible triples and then we present an efficient mechanism to sample from this distribution over the triples. Lastly, we describe a training routine that we used to effectively utilise these augmented triples in training link predictors. 

\subsection{Probabilistic Formulation}
Borrowing notation from \cite{dobrowolska2021neural}, we define \textit{Knowledge Graph} \( \mathcal{G} = \{(h, r, t)\} \subseteq \mathcal{E}\times \mathcal{R} \times \mathcal{E} \) as a set of triples of the form $(h, r, t)$ such that \( h,t \in \mathcal{E}\), \( r \in \mathcal{R}\) and $h \neq t$, ie, no self loops in the graph. It can thus be viewed as a \textit{directed graph} with \textit{head entity}(\(h\)) and \textit{tail entity}(\(t\)) as the nodes and the \textit{relation type}(\(r\)) as corresponding edge label. 

There can be many ways to factorize the distribution of the triples, one of which is as follows: 
\begin{equation}
    p(h,r,t) = p(h,t) * p(r|h,t)
\end{equation}
where \( p(h,t) \) is the distribution over the possible edges in the graph and (\( p(r|h,t)\)) is the distribution of the relations conditioned on an edge denoted by $(h, t)$. \\
Since knowledge graphs are known to be inherently sparse (statistics of some benchmarks are provided in table \ref{table:datasets}) and the entities have a certain "type" that categorizes them semantically, we further propose to model the entities in the KG as a set of \textit{clusters}, where all the generated clusters are disjoint. We thus arrive at the following factorization:
\begin{align}
    p(h,r,t) &= p(r|h,t)*\sum_{\forall cluster} p(cluster)* p(h,t|cluster) \nonumber \\
             &= p(cluster_i) * p(h,t|cluster_i) *p(r|h,t) \label{factorization}
\end{align}
where \( cluster_i\) is the cluster containing a given entity tuple \({(h,t)}\). 

\subsection{Generating Entity Clusters}
We now define the matrices: $A$ $\in \mathbb{R}^{|\mathcal{E}| \times |\mathcal{R}|}$ (call it \textit{head-relation} matrix) and $B$ $\in \mathbb{R}^{|\mathcal{E}| \times |\mathcal{R}|}$ (\textit{tail-relation} matrix) as follows:
\begin{align}
    A = [a_{ij}]_{|\mathcal{E}| \times |\mathcal{R}| } , ~~a_{ij} = |\{(i,j,k) \in \mathcal{G}\}| \label{eq: h_rMatrix} \\
    B = [b_{ij}]_{|\mathcal{E}| \times |\mathcal{R}| },  ~~b_{ij} = |\{(k,j,i) \in \mathcal{G}\}| \label{eq: t_rMatrix}
\end{align}
 
One way to generate disjoint entity clusters is by using an algorithm such as higher order spectral clustering \cite{DBLP:journals/corr/abs-2011-05080}, that can find cuts in the KG where the set of nodes in each cut form a weakly connected component. However note that it is non-trivial to achieve this clustering over the original KG directly, since there is no natural way to assign weights to the edges (where each edge is a relation type). We rather consider the digraph generated by taking the entities as nodes and the elements of the affinity matrix $C$ (eq \ref{affinity_matrix} below) as corresponding weighted edges. Intuitively speaking, matrix $C$ can be viewed as a co-occurence matrix of entity pairs marginalized over all relation types. Formally, it can be represented as follows:
\begin{align}
\label{affinity_matrix}
C = A B^{T}   
\end{align}
Although spectral clustering over $C$ can provide the desired entity clusters but despite its suitability to our problem, spectral clustering is computationally expensive with a high memory and time overhead which renders it impractical for the KGs with large number of entities. Alternately, seeking inspiration from the GloVe \cite{pennington2014glove} algorithm, we first generate lower dimensional representations of the entities by performing \textbf{Non-Negative Matrix Factorization (NNMF)} \cite{cichocki2009fast} (a brief introduction is provided in section \ref{subsec:nnmf}) of the affinity matrix $C$. In order to utilize all available information, the two matrices that are obtained from NNMF, represented as $W_1$ and $W_2$ here, are concatenated along the column dimension, denoted as $W^{'}$, and passed through a standard clustering method that operates on euclidean space to generate a partition denoted by $\mathcal{K}$ $=$ $\{K_i\}_{i=1}^{N}$, such that $\bigcup_{i=1}^{N} K_i$ $=$ $\mathcal{E}$ and $\forall i, j; i \neq j$ $\rightarrow$ $K_i \cap K_j$ $= \emptyset$. The size of $\mathcal{K}$, ie, the number of clusters is a used defined hyperparameter. We use \textit{Agglomerative clustering} \cite{Zepeda-Mendoza2013} over $W^{'}$ as it provides a reasonable tradeoff between speed and quality.

\subsection{Sampling}
\label{subsec:sampling}
To generate and subsequently augment new triples to the training set, we utilize the factorisation of $p(h,r,t)$ proposed in equation \eqref{factorization} by first estimating the distributions p($cluster_i$), $p(h,t|Cluster_i)$ and $p(r|h,t)$ from the statistics of the training data. To simplify the computation, we assign a uniform distribution to $p(cluster)$, thus the probability of selecting all clusters is equal. Similarly, we also assign a uniform distribution to all pairs of entities $(h, t)$ in a given cluster $i$. 

Lastly, to estimate $p(r|h,t)$ we use matrices $A$ and $B$. We perform element wise multiplication of the row of $A$ corresponding to head entity $h$ with the of row of $B$ corresponding to tail entity $t$. This provides us a vector $\vec{d} \in \mathbb{R}^{|\mathcal{R}|}$ which is further normalized by dividing the entry in each dimension of $\vec{d}$ by the sum of all entries in $\vec{d}$ such that it becomes a probability simplex and is then used to sample the relation type $r$, weighted by its corresponding probability value in normalised $\vec{d}$, finally giving us a new triple $(h, r, t)$.

The above procedure is repeated until we obtained a desired number of triples to augment. We denote the set of newly generated triples by $\mathbf{S}$. The complete pipeline is provided in algorithm \ref{algo}.

\begin{algorithm}[h]
\small

\SetAlgoLined
\Input{ Knowledge Graph $\mathcal{G}$, number of clusters $N$, number of triples to generate $L$}
\textbf{Initialize: } $\mathbf{S}~=~\{\}$; matrices $A$, $B$, $C$ \hfill
\Comment{ via Eq \ref{eq: h_rMatrix}, \ref{eq: t_rMatrix} and \ref{affinity_matrix} respectively} 
$W_1$, $W_2$ $\gets$ NNMF(C)  \hfill \Comment{Eq \ref{eq:do_nnmf}} 
$W' ~ \leftarrow$ [$W_1$, $W_2$]  \hfill 
\Comment{Column-wise Concatenation}   
$\mathcal{K}$   $\leftarrow$ AggClustering($W'$, $N$)  \hfill
\Comment{Agglomerative Clustering of $W'$} 
\While{|$\mathbf{S}$| < $L$}{
    Generate New Triple $(h', r', t')$, given partitioning $\mathcal{K}$ \hfill \Comment{Section \ref{subsec:sampling}} 
    $\mathbf{S}\leftarrow \mathbf{S}$ $\bigcup$ $\{(h', r', t')\}$ \hfill \Comment{Set Union} 
 }
 \Output{$~\mathbf{S}$}
 \caption{\textbf{Proposed NNMFAug Method}}
 \label{algo}
\end{algorithm}

\subsection{Routine to Monotonically Increase Augmented Data Size}
\label{subsec:routine}
Rather than augmenting the training data with the entire set $\mathbf{S}$, we follow a routine that monotonically increases the number of new triples $r$, added per epoch, as the training progresses. $r$ is calculated as $(\frac{e}{E})^{k} \times |\mathbf{S}|$, where $e$ is the current training epoch, $E$ is the total number of training epochs, $k \in \mathbb{Z}^+$ is a hyperparameter and $|\mathbf{S}|$ is the size of set $\mathbf{S}$. We empirically observed that this routine helps as the augmented triples can be sometimes noisy and gradually introducing them to the model can help the model generalize better. Further analysis for the hyperparameter $k$ has been done in section \ref{sec:analysis}.

\section{Experiments}
We evaluate the efficacy of the proposed data augmentation method on two widely used neural link prediction models: \textit{TransE} ~\cite{NIPS2013_1cecc7a7} and \textit{RotatE} ~\cite{sun2018rotate} over two widely used datasets: \textit{Wordnet18RR} (WN18RR) ~\cite{DBLP:journals/corr/DettmersMSR17} and \textit{DeepLearning50a} (DL50a) ~\cite{dataset_dl50a}. A brief introduction to neural link predictors is provided in appendix \ref{sec:link_preds}. The statistics for the datasets are provided in table \ref{table:datasets} and the evaluation protocol is briefly described in section \ref{sec:evaluation_protocol}. All the experiments have been performed using PyKg2Vec library on a single GPU with 8 GB cuda memory. 

Table \ref{table:results} report the results (averaged over 5 runs) for each of the KG model-dataset combination. The suffix "Baseline" represents the original model performance whereas "NNMFAug" shows the model performance with our augmentation strategy.

\begin{table*}[h]

    \caption{
    Results for various evaluation metrics (section \ref{sec:evaluation_protocol}). Best values are highlighted in bold.}
    \label{table:results}
    
    \centering
    \small
    \begin{tabular}{|c|c|cccccc|}
    \hline
        Dataset & Model & H@1$\uparrow$ & H@3$\uparrow$ & H@5$\uparrow$ &H@10$\uparrow$ & MRR$\uparrow$ & MR$\downarrow$ \\
        
        \hline
        
        \multirow{4}{*}{DL50a} & TransE-Baseline & 8.13 & 19.27 & 25.03 & 33.21 & 0.1597 & 506  \\
         & TransE-NNMFAug & \textbf{9.97} & \textbf{21.76} & \textbf{27.25} & \textbf{34.25} & \textbf{0.1797} & \textbf{490}  \\
        \cline{2-8}
         & RotatE-Baseline & 35.05 & 45.62 & 49.56 & 54.82 & 0.4221 & 156  \\
         & RotatE-NNMFAug & \textbf{35.62} & \textbf{46.08} & \textbf{50.10} & \textbf{55.34} & \textbf{0.4276} & \textbf{152}  \\

        \Xhline{4\arrayrulewidth}

        \multirow{4}{*}{WN18RR} & TransE-Baseline & 1.24 & \textbf{35.41} & 41.85 & 47.29 & \textbf{0.1974} & 3920  \\
         & TransE-NNMFAug & \textbf{1.31} & 35.28 & \textbf{41.99} & \textbf{47.48} & 0.1971 & \textbf{3766}  \\
        \cline{2-8}
         & RotatE-Baseline & 39.59 & 47.89 & 51.10 & 55.39 & 0.4527 & \textbf{3115}  \\
         & RotatE-NNMFAug & 39.59 & \textbf{48.09} & \textbf{51.14} & \textbf{55.67} & \textbf{0.4537} &  3195 \\
        
        \hline
        
    \end{tabular}
    
\end{table*}

\section{Analysis}
\label{sec:analysis}
In this section, we quantitavely analyze the model performance against some of the important hyperparameters of the augmentation method.

\begin{figure*}[tbp]
	\begin{subfigure}[b]{0.33\textwidth}
		\centering
		\includegraphics[width=\linewidth]{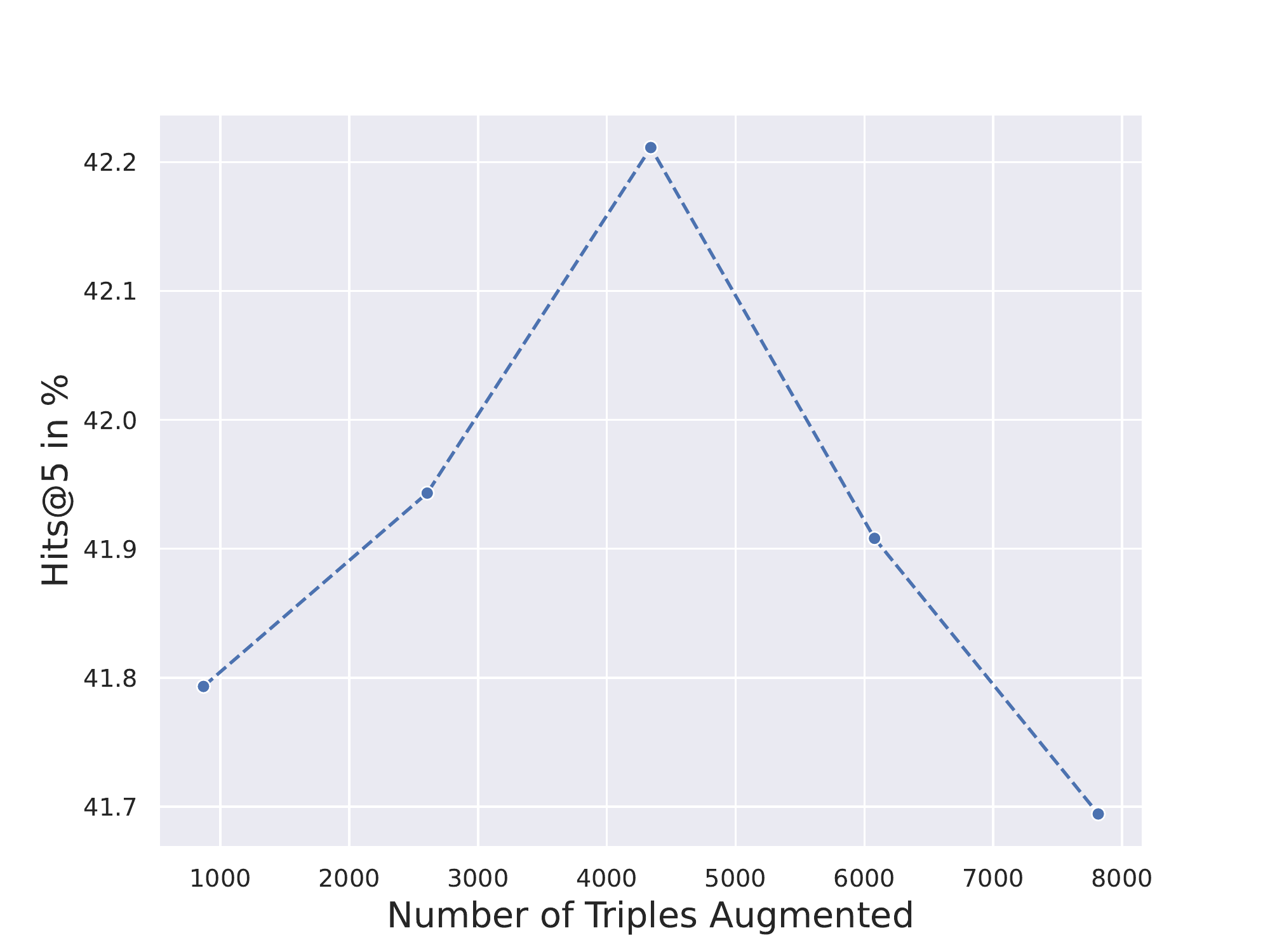}
		\caption{TransE on Wordnet18RR \\dataset.}
		\label{fig:analysis_frac}
	\end{subfigure}%
	\begin{subfigure}[b]{0.33\textwidth}
		\centering
		\includegraphics[width=\linewidth]{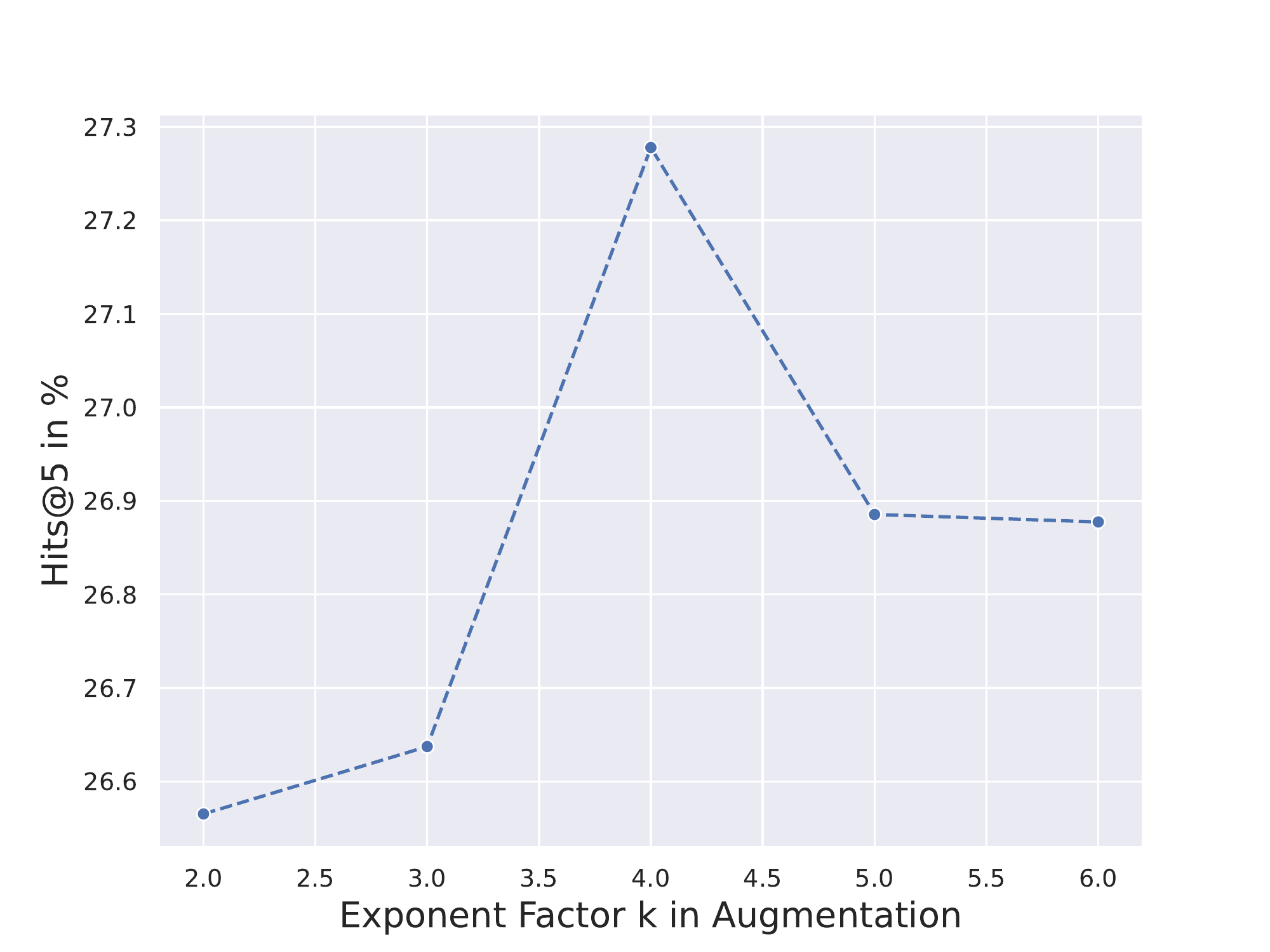}
		\caption{TransE on DL50a \\dataset.}
		\label{fig:analysis_poly}
	\end{subfigure}%
	\begin{subfigure}[b]{0.33\textwidth}
		\centering
		\includegraphics[width=\linewidth]{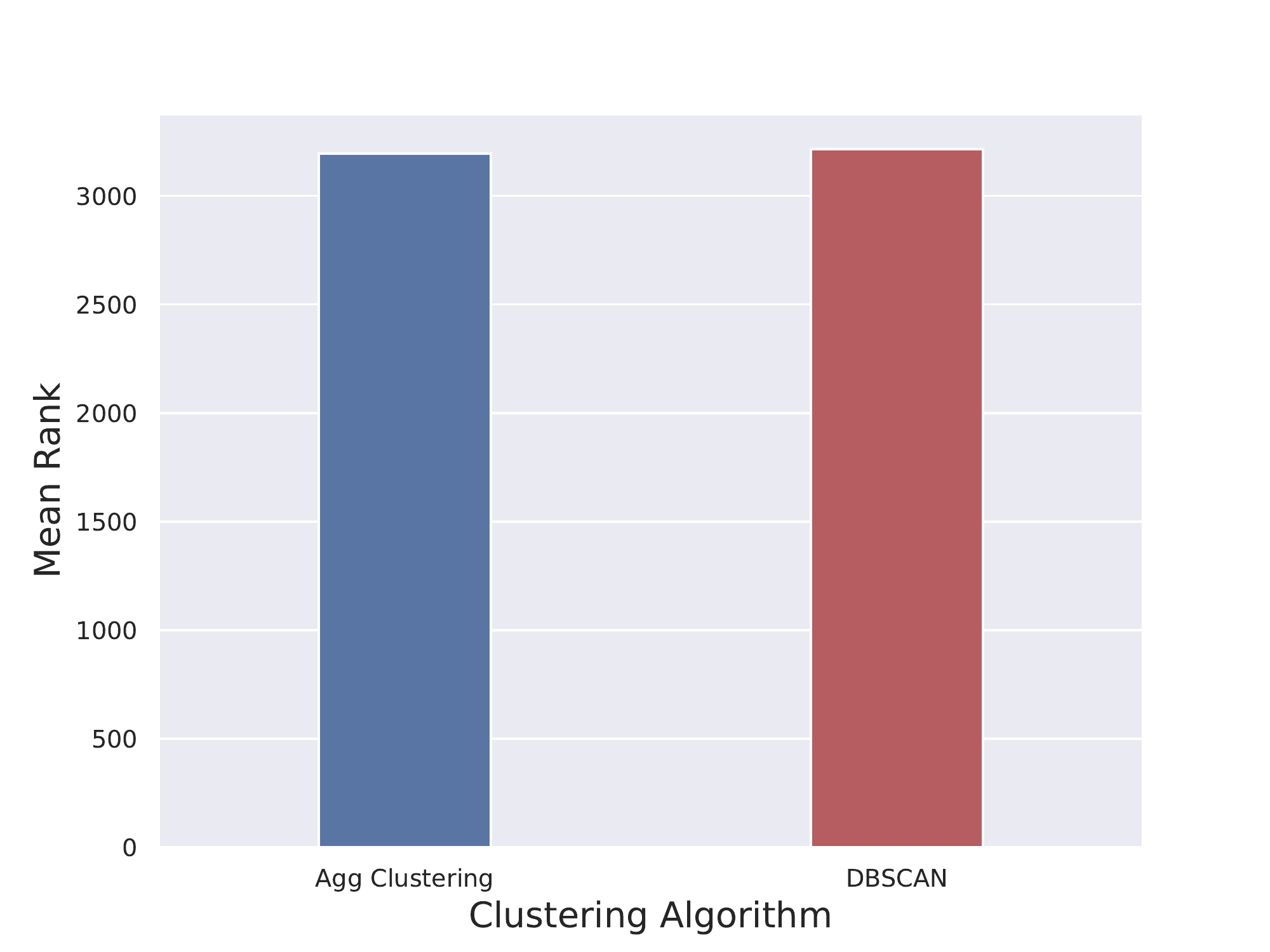}
		\caption{RotatE on Wordnet18RR \\dataset.}
		\label{fig:analysis_clus}
	\end{subfigure}%
	
	\caption{Analysis for the variation of \textit{Number of Triples Augmented} (left subfigure), \textit{Factor $k$ in Augmentation Routine} (middle subfigure) and the \textit{Clustering Algorithm Used} (right subfigure).}	
	\label{fig:analysis}
\end{figure*}

\textbf{1) Size of the Augmented data: } The number of triples augmented to the training set has a direct affect on the downstream model performance, as shown in figure \ref{fig:analysis_frac}. It is interesting to note the \textit{Inverted V-shaped} curve for the metrics. Improved downstream performance of neural networks via more augmented data is a well-known phenomenon \cite{wei2019eda, conv_aug}, however, in our case it is evident that augmentation beyond a certain point can affect the models negatively. We hypothesize that this is due to presence of some \textit{false positive triples}, which are generated as a consequence of sampling (a probabilistic procedure), that can hinder the learning process because of excess noise. We also point that the location of the peak of the curve can vary depending upon the dataset.

\textbf{2) Exponent factor in Augmentation Routine: } The exponent $k$ (section \ref{subsec:routine}) that governs the number of generated triples augmented to the training data per epoch, also has a direct impact on the downstream performance, as shown in figure \ref{fig:analysis_poly}. Here as well, we observe that gradually increasing the factor first improves the metrics to a peak value, post which the performance decreases, showing a similar \textit{Inverted V-shaped} curve trend. It follows a similar reasoning as previous subsection that augmenting the training data with larger number of triples in the early phases of training can hinder the learning due to presence of some false positive triples. Thus, its important to learn from original training data in the early phases and follow a monotonic increment routine in augmentation.

\textbf{3) Clustering Algorithm: } We evaluate the performance of the models against the clustering algorithm used to group entities into multiple clusters. We compare the performance of two widely known clustering algorithms: \textit{Agglomerative Clustering} \cite{Zepeda-Mendoza2013} (used in this work) and \textit{DBSCAN} \cite{dbscan_clus}. From the comparison shown in figure \ref{fig:analysis_clus}, its evident that our strategy is \textit{less succeptible} to the clustering algorithm used and thus has a wide applicability.

\section{Conclusion}
In this work, we presented a novel data augmentation strategy for the task of link prediction in Knowledge Graphs named NNMFAug, which is agnostic to a specfic method and dataset as well as capable of performing data augmentation efficiently in an offline manner while utilizing multiprocessing. We show that NNMFAug provides consistent gains over multiple dataset and model combinations as well as anticipate that this work will draw attention and also pave way for more probabilistic as well as deterministic methodologies in this understudied space.

\bibliographystyle{plainnat}
\bibliography{neurips_2021}

\newpage
\appendix

\section{Appendix}

\begin{table*}[h]
    \caption{Dataset statistics}
    \label{table:datasets}
    
	\centering\small
	\begin{tabular}{|l|cc|cccc|}
		\hline
		&               &              & \multicolumn{4}{|c|}{ \textbf{\# Triples}} \\ 
		\cline{4-7}
		\textbf{Dataset}   &  \textbf{\# Entities}  & \textbf{\# Relations}  &  Training  & Validation  & Test & Total \\ 
		\hline
		\emph{WN18RR}    & 40,943 & 11  & 86,835   & 3034   & 3134   & 93,003 \\ 
		\emph{DL50a} & 2705 & 20  & 6000  & 770    & 1249 & 8019  \\     
		\hline
	\end{tabular}

\end{table*}

\subsection{Neural Link Predictors}
\label{sec:link_preds}
\textit{Link Predictors} (in the KG setting) are models trained to maximize the likelihood of the triples in training data and further used to assign likelihood of new triples being correct during inference time (more details in section \ref{sec:evaluation_protocol}). Neural Link predictors can thus be seen as deep learning based link predictors which essentially learn low dimensional representations for the entities (represented by  $\mathbf{E}^{|\mathcal{E}| \times d}$) and relations (represented by $\mathbf{R}^{|\mathcal{R}| \times d}$) in the KG, along with some other trainable parameters (represented by $\mathbf{\theta}$), through back-propagation\cite{Nickel_2016}. With these trainable parameters, a neural link predictor defines a \textbf{scoring function} $f$ (mostly heuristic) over the embedding vectors $\mathbf{h}$, $\mathbf{r}$, $\mathbf{t}$ of an input triple $(h, r, t)$ that are indexed from $\mathbf{E}$ and $\mathbf{R}$ respectively such that $f(\mathbf{h}, \mathbf{r}, \mathbf{t}; \mathbf{\theta}) \rightarrow \mathbb{R}$ ; assigns a likelihood to the triple being correct. 
While there is a rich literature surrounding the designs of these link predictors, in this work we have focused on two popular models: \textit{TransE} and \textit{RotatE}. Their scoring functions are provided in table \ref{table:score_function}. Note that the $d$ dimensional embedding space can be \textit{real valued} (see TransE in table \ref{table:score_function}) or \textit{complex valued} (see RotatE in table \ref{table:score_function}).

It is also noteworthy that along with the positive triples provided in the training data, these models are fed a large number of \textit{negative} or \textit{corrupt} triples generated by the same mechanism as described in section \ref{sec:evaluation_protocol} so that the models can learn to distinguish correct triples from the incorrect ones. This mechanism is usually referred as \textit{negative sampling} and is intuitively same as the \textit{negative sampling} procedure of \textit{Word2Vec} algorithm \cite{DBLP:journals/corr/MikolovSCCD13}.
\\


\begin{table*}[h]

    \caption{The scoring functions $ f (h,r,t)$ and embedding constraints of \textit{TransE} and \textit{RotatE} models. $\mathbb{C}$ represents the complex space and $d$ denotes the embedding dimensions.}
    \label{table:score_function}
    
    \centering
    \small
    \begin{tabular}{|c|c|c|}
    \hline
    Model & Score function & Embedding Constraints\\
    \hline
    TransE & $-\norm{\textbf{h}+\textbf{r}-\textbf{t}}$ & $\textbf{h},\textbf{r},\textbf{t} \in \mathbb{R}^d$\\
    \hline
    RotateE & $-\norm{\textbf{h}\circ\textbf{r}-\textbf{t}}^2$ & $\textbf{h},\textbf{r},\textbf{t} \in \mathbb{C}^d, \norm{\textbf{r}_i}  = 1$\\
    \hline\end{tabular}
\end{table*}

\subsection{Evaluation Protocol}
\label{sec:evaluation_protocol}

For a given triple $(h, r, t)$ in the test set, either the head entity $h$ or the tail entity $t$ is assumed to be missing and the aim is to predict the missing entity given the relation and the other entity. Without of loss of generality, lets assume that $h$ is missing. First, a set of $\mathcal{E} - 1$ corrupt triples is generated by appending each entity $e \in \mathcal{E} \textbackslash h$ to $(r, t)$, generating a total of $\mathcal{E}$ triples, including the original correct test triple. These triples are then passed through the neural link predictor and subsequently sorted in descending order of the scores. We then obtain the rank of correct triple $(h, r, t)$. The same procedure is repeated for both the head and tail entities across the entire test set and the results are averaged to finally report the Mean Reciprocal Rank (MRR), Mean Rank (MR) and the percentage of correct triples in the top $R$ ranks (Hits@R) for $R = 1, 3, 5$ and $10$, after being sorted. For Mean Reciprocal Rank as well as the Hits@R metrics higher values are better whereas for Mean Rank lower is better.

\subsection{Non Negative Matrix Factorization}
\label{subsec:nnmf}
Non-Negative Matrix Factorization is a decomposition of a matrix $M^{m \times n}$ into 2 component matrices $W_1^{m \times p}$ and $W_2^{p \times n}$ such that 
\begin{equation}
    M = W_1 W_2
\end{equation}
with the constraint that all the three Matrices have \textbf{non-negative elements}. While the factorization is not necessarily unique, polynomial closed form solutions can be calculated by enforcing additional constraints on $W_1$ and $W_2$ matrices. However, in practice approximate methods prove to be a Time and Memory efficient alternative. In this work, we have used the NNMF implementation provided by scikit-learn library \footnote{https://scikit-learn.org/stable/modules/generated/sklearn.decomposition.NMF.html} that uses alternating minimization of $W_1$ and $W_2$ to minimize the objective function ($\mathcal{L}(W_1,W_2,M)$) below:
\begin{align}
\label{eq:do_nnmf}
    \mathcal{L}(W_1,W_2,M) &= 0.5 * \norm{M - W_1W_2}^2_{fro} + \alpha*\Omega(W_1,W_2)
\end{align}
where, $fro$ represents the \textit{frobenius} norm of the matrix, $\alpha$ is a hyper-parameter  and $\Omega(W_1,W_2)$ is the regularization term such that:
\begin{align}
    \Omega(W_1,W_2) = c *(\norm{vec(W_1)}_1 +\norm{vec(W_2)}_1) +0.5*(1-c) * (\norm{W_1}^2_{fro} + \norm{W_2}^2_{fro})
\end{align}
where $c$ is another hyper-parameter and $\norm{}_1$ is the L1 norm.

Time complexity of this algorithm is $\mathcal{O}(mpn \times q)$, where $q$ is the number of iterations performed during alternate minimization.

\subsection{Related Work}
There have been a few prior works performing \textit{data augmentation} for link prediction in knowledge graphs, however it still remains a fairly new and explored research area. \cite{pmlr-v80-lacroix18a} introduced the concept of augmenting the training data by adding new triples consisting of \textit{inverse relations} which improves the performance of neural link predictors over multiple benchmarks. In another work, \cite{DBLP:journals/corr/MinerviniDRR17} proposed a method to generate sets of adversarial examples that maximizes an inconsistency loss which encodes specific background knowledge. In a more recent work, \cite{dobrowolska2021neural} revisit the notion of learning novel concepts in Knowledge graphs in a more principled way. More succinctly, they propose a method to cluster the entities where each cluster represents a concept. There are fundamental differences between our work and theirs in that: (i) they introduce a new "relation type" per cluster to generate new triples whereas we utilize the existing set of entities and relation types to generate new triples in a probabilistic manner; (ii) our method follows along the lines of Glove \cite{pennington2014glove} algorithm since we seek latent entity embedding vectors via NNMF where the loss is minimized based on co-occurence of entity-relation pairs while accounting for the edge direction.


\end{document}